\documentclass[10pt,twocolumn,letterpaper]{article}

\usepackage{cvpr}
\usepackage{times}
\usepackage{epsfig}
\usepackage{graphicx}
\usepackage{amsmath}
\usepackage{amssymb}

\newcommand\remove[1]{}

\usepackage[breaklinks=true,bookmarks=false]{hyperref}

\cvprfinalcopy 

\def\cvprPaperID{3468} 

\ifcvprfinal\fi
\begin{document}

\title{Cross-Classification Clustering: An Efficient Multi-Object Tracking Technique for 3-D Instance Segmentation in Connectomics}

\author{Yaron Meirovitch$^{1,2}$\thanks{\mbox{These authors equally contributed to this work} {\tt \{yaronm,lumi\}@mit.edu}}\ , Lu Mi$^1$\footnotemark[1]\ , Hayk Saribekyan$^{1}$, Alexander Matveev$^{3}$, David Rolnick$^{1}$, Nir Shavit$^{1,4}$ \\
$^1$MIT, \mbox{   } $^2$Harvard University, \mbox{   } $^3$Neural Magic Inc., \mbox{   } $^4$Tel-Aviv University \\
}



\maketitle


\begin{abstract} 
Pixel-accurate tracking of objects is a key element in many computer vision applications, often solved by iterated individual object tracking or instance segmentation followed by object matching. Here we introduce {\it cross-classification clustering} (3C), a technique that simultaneously tracks complex, interrelated objects in an image stack. The key idea in cross-classification is to efficiently turn a clustering problem into a classification problem by running a logarithmic number of independent classifications per image, letting the cross-labeling of these classifications uniquely classify each pixel to the object labels. We apply the 3C mechanism to achieve state-of-the-art accuracy in connectomics -- the nanoscale mapping of neural tissue from electron microscopy volumes. Our reconstruction system increases scalability by an order of magnitude over existing single-object tracking methods (such as flood-filling networks). This scalability is important for the deployment of connectomics pipelines, since currently the best performing techniques require computing infrastructures that are beyond the reach of most laboratories. Our algorithm may offer benefits in other domains that require pixel-accurate tracking of multiple objects, such as segmentation of videos and medical imagery.
\end{abstract}

\section{Introduction}
\label{sec:introduction}

\begin{figure}
\centering 
	 \includegraphics[width=1\linewidth]{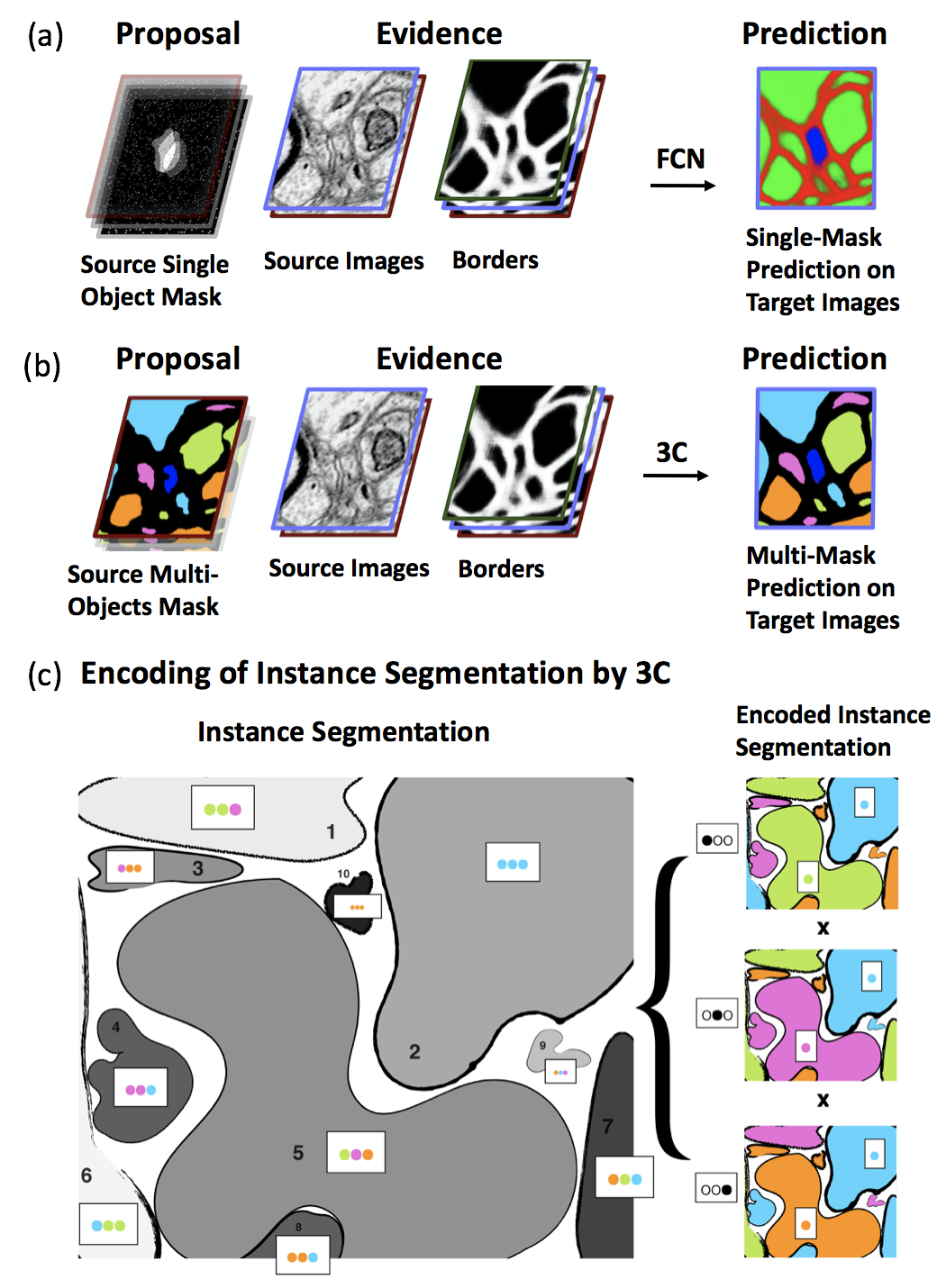} 
	 \caption{
	 (a) Single-object tracking using flood-filling networks~\cite{FFN}, 
	 (b) Multiple-object tracking using our cross-classification clustering (3C) algorithm, 
	 (c) The combinatorial encoding of an instance segmentation by 3C. One segmented image with 10 objects is encoded using three images, each with 4 object classes.}	
	 \label{Fig: introduction and Relabeling for Cross Classification}
\end{figure}

Object tracking is an important and extensively studied component in many computer vision applications ~\cite{adam2006robust,fan2010human,girdhar2018detect,held2016learning,kang2016object,wang2017large,yoo2017action,zhang2017multi}. It occurs both in video segmentation
and in 3-D object reconstruction based on 2-D images. Less attention has been given to efficient algorithms performing simultaneous tracking of multiple interrelated objects~\cite{girdhar2018detect} in order to eliminate the redundancies of tracking multiple objects via repeated use of single-object tracking. This problem is relevant to applications in medical imaging \cite{dou20173d,drozdzal2018learning,kalinin20183d,lahiri2017generative,lee2017deep,milletari2016v} as well as videos \cite{fagot2016improving,perera2006multi,xiang2015learning,zhang2008global}.



The field of connectomics, the mapping of neural tissue at the level of individual neurons and the synapses between them, offers one of the most challenging settings for testing algorithms to track multiple complex objects.
Such synaptic level maps can be made only from  high-resolution images taken by electron microscopes, where the sheer volume of data that needs to be processed (petabyte-size image stacks), the desired accuracy and speed (terabytes per hour \cite{Lichtman-Pfister-Shavit-Survey}), and the  complexity of the neurons' morphology, present a daunting computational task. By analogy to traditional object tracking, imagine that instead of tracking a single sheep through multiple video frames, one must track an entire flock of sheep that intermingle as they move, change shape,  disappear and reappear, and obscure each other \cite{LichtmanDenk}. 

As a consequence of this complexity, several highly successful tracking approaches from other domains, such as the ``detect and track'' approach \cite{girdhar2018detect}, are less immediately applicable to connectomics. 

Certain salient aspects are unique to the connectomics domain:
%
{\bf a)} All objects are of the same type (biological cells); sub-categorizing them is difficult and has little relevance to the segmentation problem. 
{\bf b)} Most of the image is foreground, with tens to hundreds of objects in a single megapixel image.
{\bf c)} Objects 
have intricate, finely branched shapes and no two are the same. 
{\bf d)} Stitching and alignment of images can be imperfect, and the distance between images ($z$-resolution) is often greater than between pixels of the same image ($xy$-resolution), sometimes breaking the objects' continuity.
{\bf e)} Some 3-D objects are laid out parallel to the image stack, spanning few images in the $z$ direction and going back and forth in that limited space with extremely large extensions in some image planes.

\begin{figure}
\centering
	 \includegraphics[width=1\linewidth]{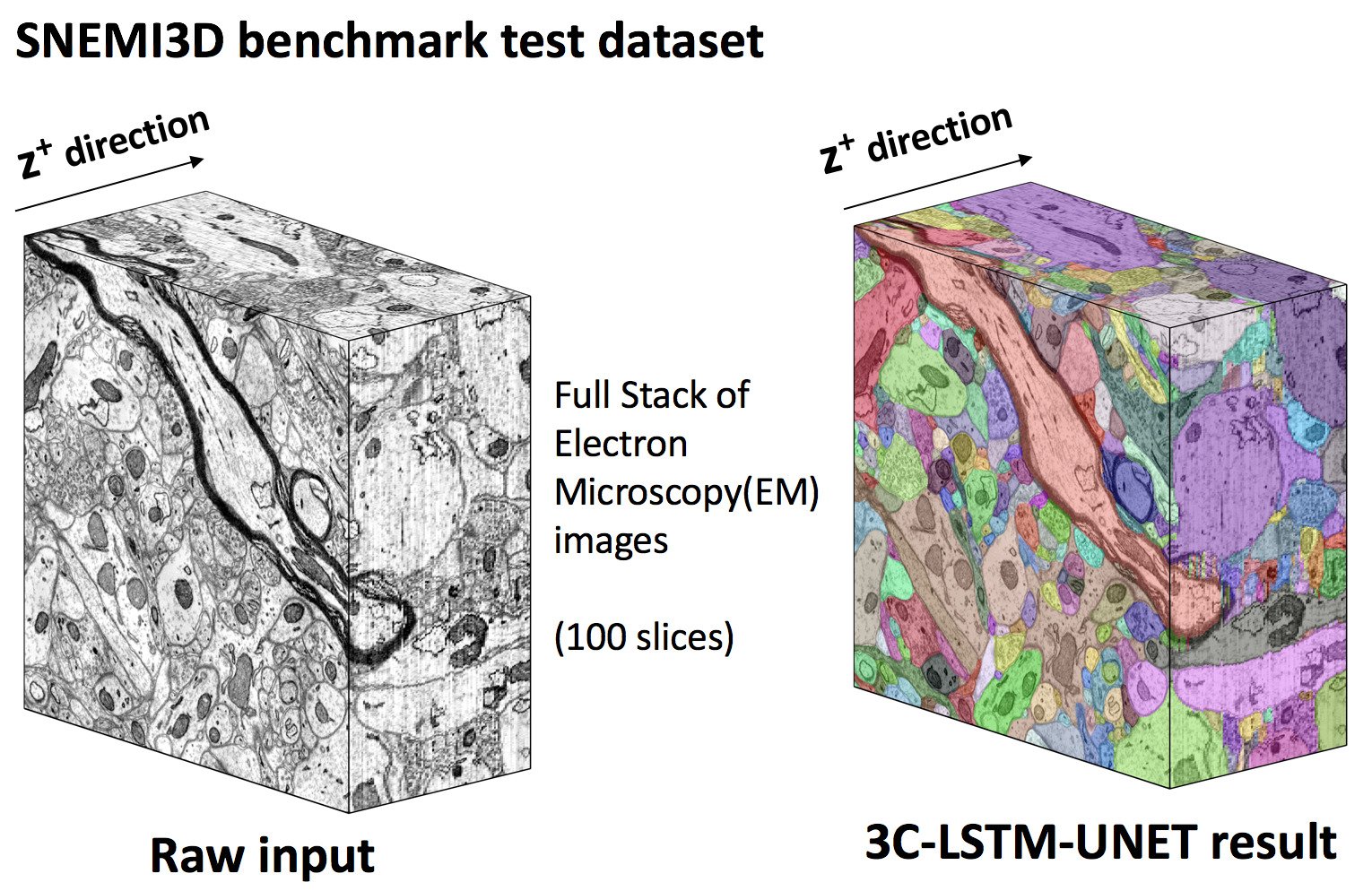} 
	 \caption{The raw input electron microscopy (EM) full image stack and our 3C-LSTM-UNET results in the SNEMI3D benchmark.}	
	 \label{Fig: 3dcube3c}
\end{figure}

In this work, we introduce 3C, a technique that achieves volumetric instance segmentation by transferring segmentation knowledge from one image to another, simultaneously classifying the pixels of the target image(s) with the labels of the matching objects from the source image(s). 
This algorithm is optimized for the setting of connectomics, in which objects frequently branch and come together, but is suitable for a wide range of video-segmentation and medical imaging applications. 


The main advantage of our solution is its ability, unlike prior single-object tracking methods for connectomics~\cite{FFN,meirovitch2016multi}, to simultaneously and jointly segment neighboring, intermingled objects, thereby avoiding redundant computation. In addition, instead of extending single masks, our detectors perform clustering by taking into account information on all visible objects. 

The efficiency and accuracy of 3C are demonstrated on four connectomics datasets: the public SNEMI3D benchmark dataset, shown in Figure~\ref{Fig: 3dcube3c}, the widely studied mouse somatosensory cortex dataset~\cite{kasthuri2015saturated} ({\it S1}), a Lichtman Lab dataset of the V1 region of the rat brain ({\it ECS}), and a newly aligned mouse peripheral nervous system dataset ({\it PNS}), where possible, comparing to other competitive results in the field of connectomics.




\subsection{Related Work}
A variety of techniques from the past decade have addressed the task of neuron segmentation from electron microscopy volumes. An increasing effort has been dedicated to the 
problem of densely segmenting all pixels of a volume according to foreground object instances (nerve and support cells), known as {\it saturated reconstruction}. Note that unlike everyday images, a typical megapixel electron microscopy image may contain hundreds of object instances, with very little background ($<$10\%). Below, we briefly survey the saturated reconstruction pipelines that seem to us most influential and related to the approach undertaken here. 

Algorithms for saturated reconstruction of connectomics data have proved most accurate when they combine many different machine learning techniques~\cite{beier2017multicut,lee2017superhuman}. Many of these techniques use the hierarchical approach of Andres et al.~\cite{andres2008segmentation} that employs  the well-known hierarchical image segmentation framework~\cite{arbelaez2011contour,haris1998hybrid,najman1996geodesic,ren2003learning}.
This is still the most common approach in connectomics segmentation pipelines: first detecting object borders in \mbox{2-D}/\mbox{3-D} and then gradually agglomerating information to form the final objects~\cite{beier2017multicut,ciresan2012deep,jain2007supervised,knowles2016rhoananet,lee2017superhuman,brianHarvardCVPR2019,Matveev2017Multicore,turaga2010convolutional}. The elevation maps obtained from the border detectors are treated as estimators of the true border probabilities~\cite{ciresan2012deep}, which are used to define an over-segmentation of the image, foreground connected components on top of a background canvas. The assumption is that each of the connected components straddles at most a single true object. Therefore it may need to be agglomerated with other connected components (heuristically~\cite{helmstaedter2013cellular,helmstaedter2013connectomic,kim2014space} or based on learned weights of hand-crafted features~\cite{ andres2008segmentation,knowles2016rhoananet,nunez2013machine,parag2015context}), but it should not be broken down into smaller segments. Numerous 3-D reconstruction systems follow this bottom-up design~\cite{beier2017multicut,berning2015segem,knowles2016rhoananet,lee2017superhuman,Matveev2017Multicore,nunez2013machine,parag2015context,PlazaBerg2016}. 
A heavily engineered implementation of hierarchical segmentation \cite{lee2017superhuman} still occupies the leading entry in the (still active) classical SNEMI3D connectomics contest of 2013~\cite{arganda2015crowdsourcing}, evaluated in terms of the uniform instance segmentation correctness metrics (normalized Rand-Error~\cite{unnikrishnan2007toward} and  Variation of Information~\cite{meilua2007comparing}).  

%

A promising new approach was recently taken with the introduction of flood-filling networks (FFN; \cite{FFN}) by Januszewski et al.~and concurrently and independently of MaskExtend \cite{meirovitch2016multi} by Meirovitch et~al. As seen in \mbox{Figure}~\mbox{\ref{Fig: introduction and Relabeling for Cross Classification}(a)}, these algorithms take a mask defining the object prediction on a source image(s), and then use a fully convolutional network (FCN) to classify which pixels in the target image(s) belong to the singly masked object of the source image(s). This process is repeated throughout the image stack in different directions, segmenting and tracking a single object each time, while gradually filling the 3-D shape of complex objects. This provides accuracy improvements on several benchmarks and potentially tracks objects for longer distances \cite{FFN} compared to previous hierarchical segmentation algorithms (e.g., \cite{beier2017multicut}). 
However, these single-object trackers are not readily deployable for large-scale applications, especially when objects are tightly packed and intermingled with each other, because then individual tracking becomes highly redundant, forcing the algorithm to revisit pixels of related image contexts many times\footnote{Such approaches thus take time linear in the number of objects and in the number of pixels, with a large constant that depends on the object density.}. Furthermore, the existing single-object detectors in connectomics \cite{FFN,meirovitch2016multi} and in other biomedical domains (e.g. \cite{arbelle2018probabilistic,bise2011automatic,huebsch2015automated,rizk2014segmentation}) do not take advantage of the multi-object scene to better understand the spatial correlation between different 3-D objects. The approach taken here generalizes the single-object approach in connectomics to achieve simpler and more effective instance segmentation of the entire volume.


\subsection{Contribution}

 
We provides a scalable 
framework for 3-D instance segmentation and multi-object tracking applications, with the following contributions:
\begin{itemize}
\vspace{-.1in}
\item We propose a simple FCN approach, tackling the less studied problem of mapping an instance segmentation between two related images. Our algorithm jointly predicts the shapes of several objects partially observed in the input. 

\vspace{-.1in}
\item We propose a 
novel technique that turns a clustering problem into a classification problem by running a logarithmic number of independent classifications on the pixels of an image with $N$ objects (for possibly large $N$, bounded only by the number of pixels).
\vspace{-.1in}
\item We show empirically that the simultaneous tracking ability of our algorithm is more efficient than independently tracking all objects. 
\vspace{-.1in}
\item We conduct extensive experimentation with four connectomics datasets, under different evaluation criteria and a performance analysis, to show the efficacy and efficiency of our technique on the problem of neuronal reconstruction.
\end{itemize}




%



\section{Methodology} 
\label{sec:methodology}
We present {\it cross-classification clustering} (henceforth {\it 3C}), a technique that extends single object classification approaches, simultaneously and efficiently classifying all objects in a given image based on a proposed instance segmentation of a context-related image. One can think of the context-related image and its segmentation as a collection of labeled masks to be simultaneously remapped together to the new target image, as in Figure~\ref{Fig: introduction and Relabeling for Cross Classification}(b). The immediate difficulty of such simultaneous settings is that this generalization is a clustering problem: unlike FFNs and MaskExtend (shown in Figure~\ref{Fig: introduction and Relabeling for Cross Classification}(a)), that produce  a binary output (``YES'' for extending the object and otherwise ``NO''), in any volume, we really do not know how many classification labels we might need to capture all the objects, or more importantly how to represent those instances in ways usable for supervised learning. Overcoming this difficulty is a key contribution of 3C. 

\begin{figure}
\centering
	 \includegraphics[width=1.0\linewidth]{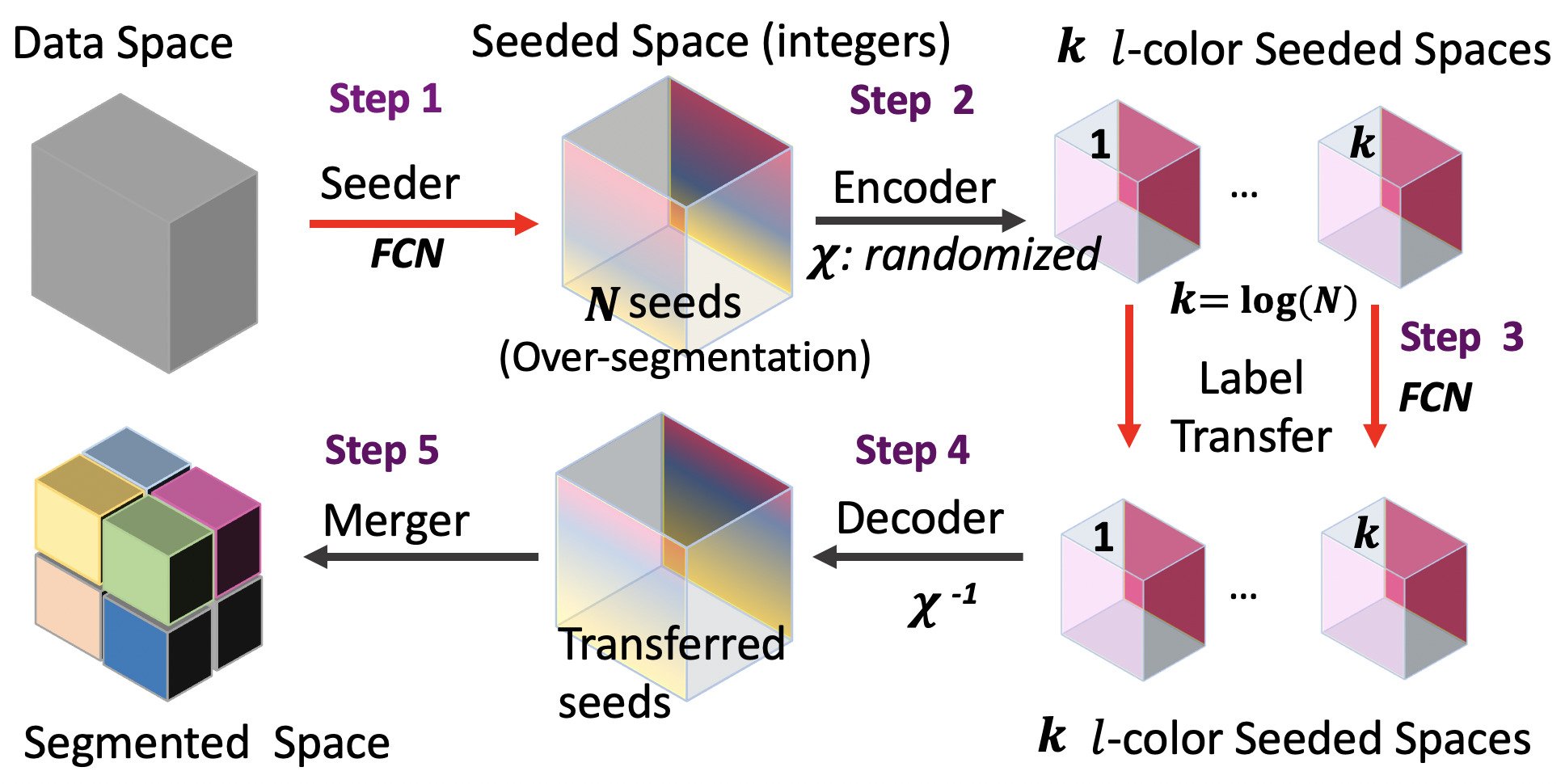} %
	 \caption{A high level view of our 3-D instance segmentation pipeline. }
	 \label{fig:TopView}
\end{figure}
 
\textbf{Cross-Classification Clustering:} We begin by explaining the main idea behind 
3C and differentiating it from single-object methods such a FFNs. We then provide a top-down sketch of our pipeline and describe how it can be adapted to other domains.

Our goal is to extend a single-object classification from one image to the next so as to simultaneously classify pixels for an \emph{a priori} unknown set of object labels. More formally, suppose that we have images $X_{prev}$ and $X_{next}$, where $X_{prev}$ has been segmented and $X_{next}$ must be segmented consistent with $X_{prev}$. Given two such images, we seek a classification function $f$ that takes as input a voxel $v$ of $X_{next}$ and a segmentation $s$ of $X_{prev}$ (an integer matrix 
representing object labels) and outputs a decision label. The function $f$ outputs a label if and only if $v$ belongs to the object with that label in $s$. If $s$ is allowed to be an over-segmentation (i.e., several labels representing the same object) then the output of $f$ should be one of the compatible labels.


For simplicity, let us assume that the input segmentation $s$ has entries from the label set $\{1,\cdots,N\}.$ We define a new space of labels, the length-$k$ strings over a predetermined alphabet $A$ (here represented by colors), where $|A|=l$ and $n{=}|A|^k \geq N$ is an upper bound on the number of objects we expect in a classification. We use an arbitrary encoding function, $\chi$, that maps labels in $\{1,\cdots,N\}$ to distinct random strings over $A$ of length $k$. In the example in Figure~\ref{Fig: introduction and Relabeling for Cross Classification}(c), $A$ is represented by $l{=}4$ colors, and $k{=}3$, so we have a total of $4^3{=}64$ possible strings of length $3$ to which the $N{=}10$ objects can be mapped.
Thus, for example, object 5 is mapped to the string (Green, Purple, Orange) and object 1 is (Green, Green, Purple). We can define the classification function $f$ on string labels as the product of $k$ traditional classifications, each with an input segmentation of labels in $A$, and an output of labels in $A$. Slightly abusing notation, let the direct product of images $\chi(s) =  \chi_1(s) \times \cdots \times \chi_k(s)$ 
be the relabeling of the segmentation $s$ where each image (or tensor) $\chi_i(s)$ is the projection of $\chi(s)$ in the $i$-th location (a single coloring of the segmentation) and $\times$ is the concatenation operation on labels in $A$. Then we can re-define $f$ on $\chi(s)$ as 
\begin{eqnarray}
\label{eq:encoding}
 f(v,\chi(s)) = f'(v,\chi_{1}(s)) \times \cdots \times f'(v,\chi_{k}(s)),  
\end{eqnarray}   
where each $f'(\chi_{k}(s))$ is a traditional classification function. The key idea is that $f'$ is a classification of $v$ based on an instance segmentation with $l$ predetermined labels. In the example in Figure~\ref{Fig: introduction and Relabeling for Cross Classification}(c), even though in the map representing the most significant digit of the original objects 5 and 1, they are both Green, when we perform the classification and take the cross labeling of all three maps, the two objects are classified into distinct labels. 
\begin{figure}
\centering
	 \includegraphics[width=1.0\linewidth]{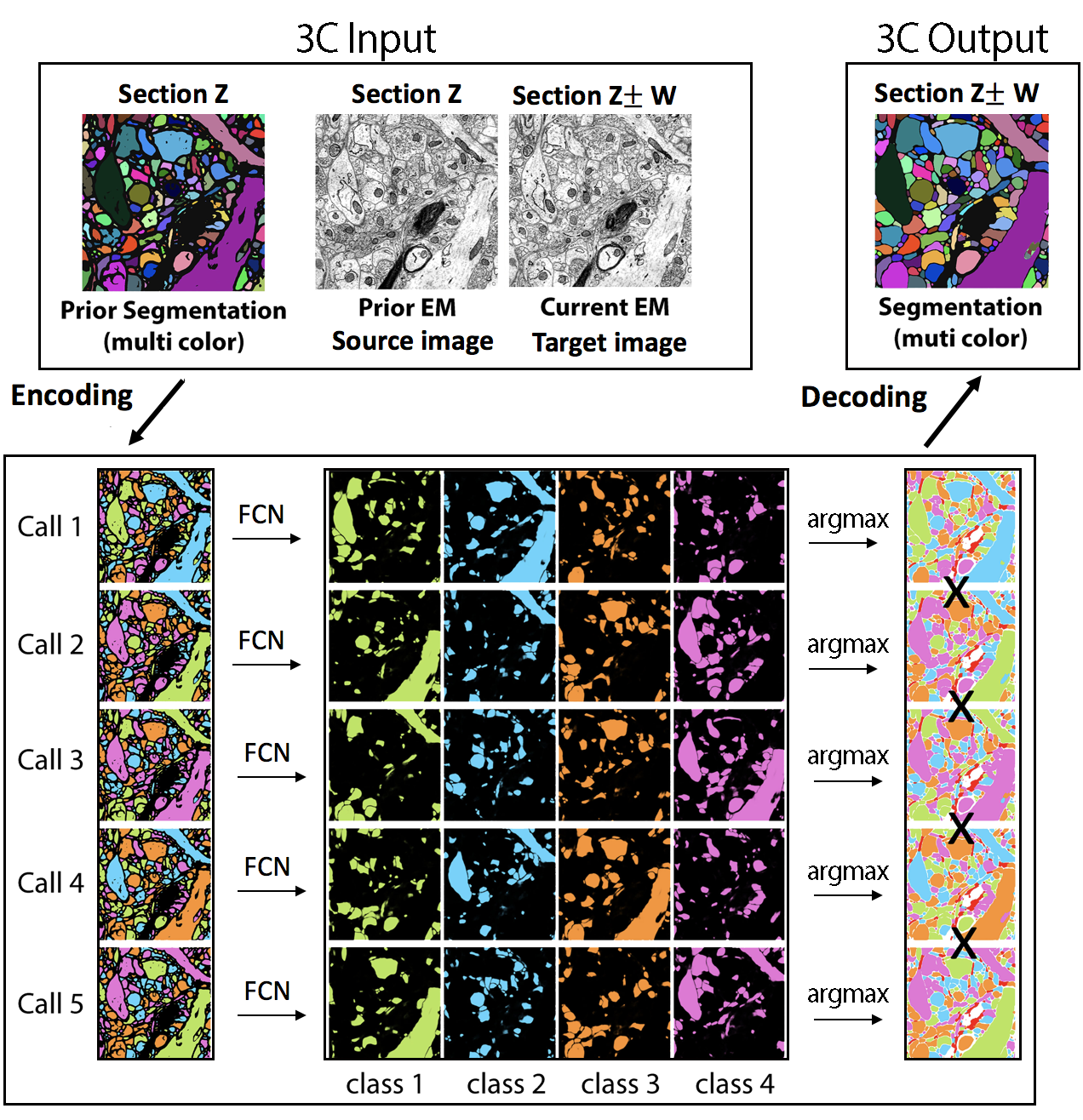} %
	 \caption{The instance segmentation transfer mechanism of 3C: Encoding the seeded image as $k$ $l$-color images using the encoding rule $\chi$ ($k{=}\log(N)$; here $k{=}5$ and $l{=}4$). Applying a fully convolutional network $k$ times    
	 to transfer each of the seed images to a respective target. Decoding the set of $k$ predicted seed images using the decoding rule $\chi^{-1}$.} %
%
	 \label{fig:relabeling_for_Cross_Classification}
\end{figure}
\begin{figure}
\centering
	 \includegraphics[width=1\linewidth]{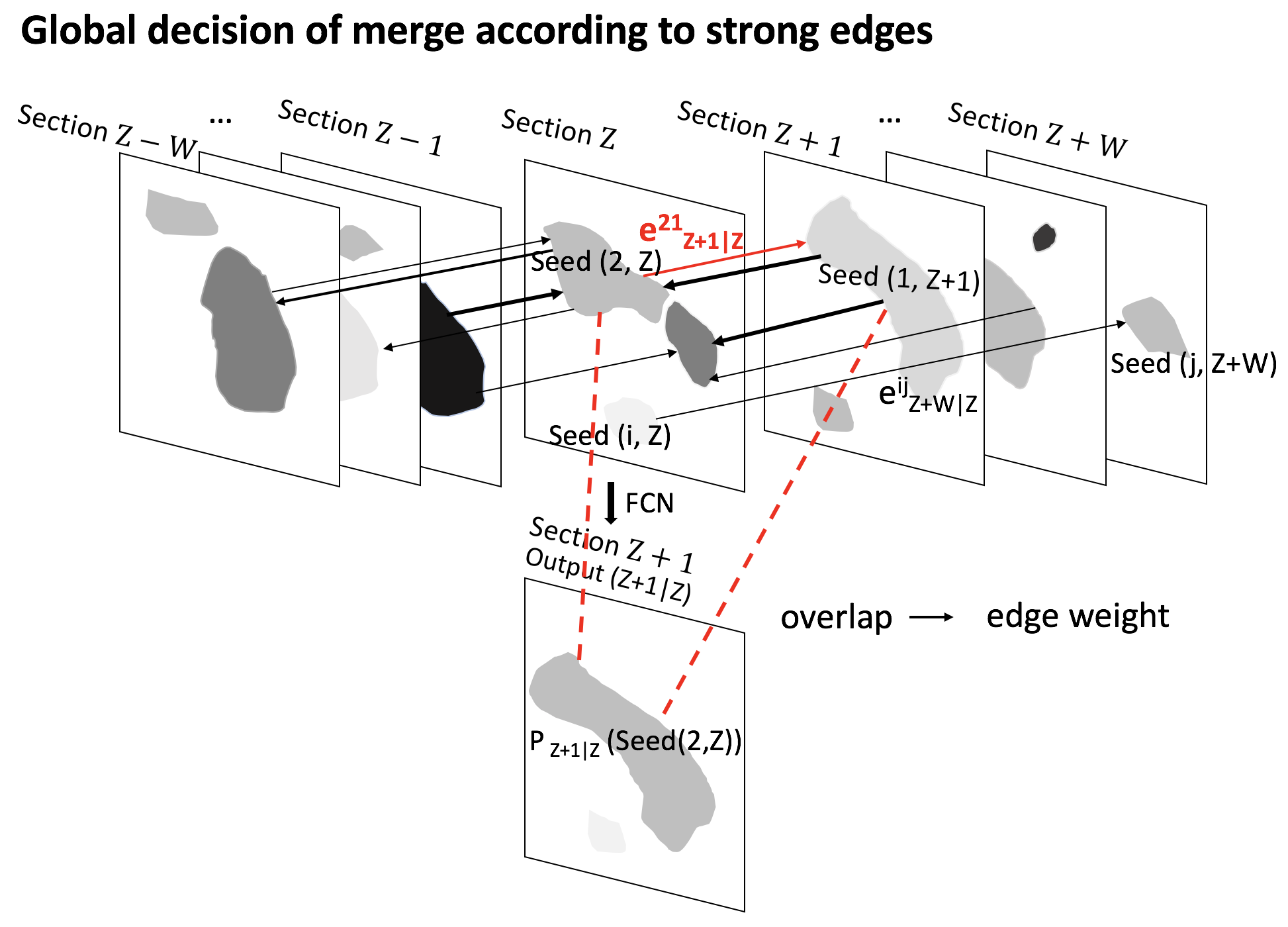} 
	 \caption{A schematic view of a global merge decision. The edge weight between seed $i$ and seed $j$ at sections $Z$ and $Z{+}W$, respectively. $e^{ij}_{Z+W|Z}$ is calculated by the ratio of the overlapping areas of seed $j$ and the 3C prediction of seed $i$ from images $Z$ to $Z{+}W$.
	 %
	 Seeds that over-segment a common object tend to get merged due to a path of strong edges. 
	 %
	 }	
	 \label{fig:global edge merge}
\end{figure}
\begin{figure*}[t]
\includegraphics[width=1.0\linewidth]{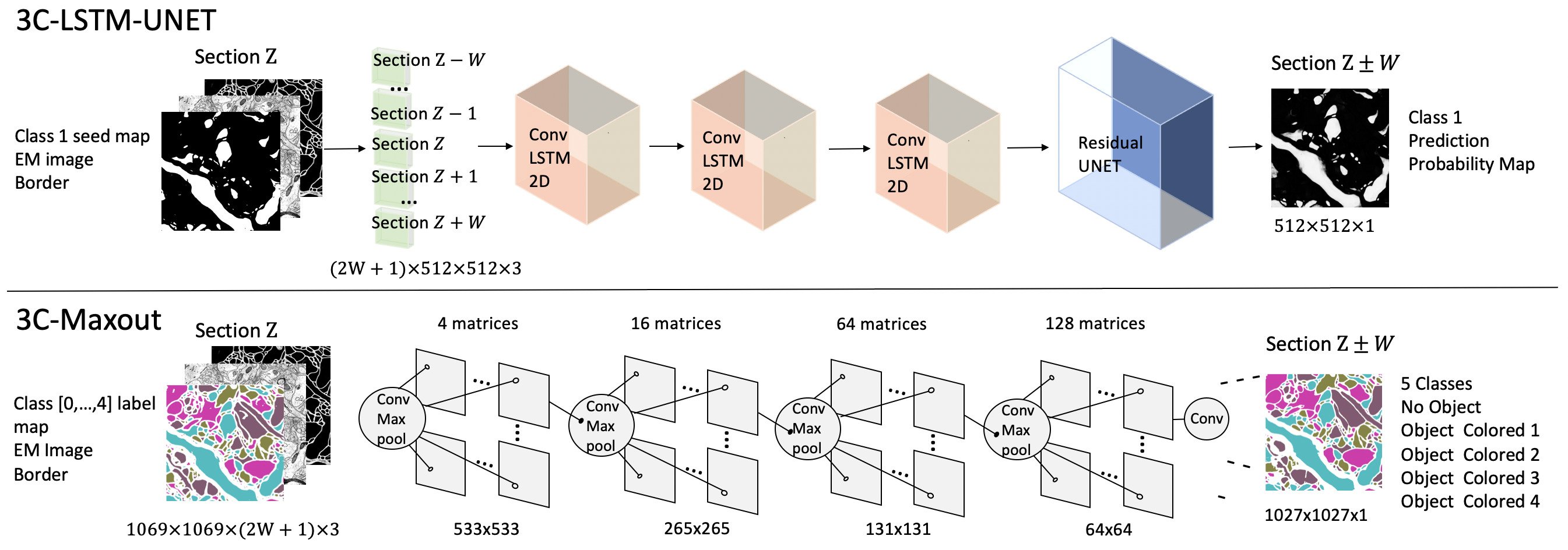}
\caption{A schematic view of the 3C networks. The input layers have $3$ channels of the raw image, seed mask and border probability, for $2W{+}1$ consecutive sections (images). The output is a feature map of seed predictions in section $Z\pm W$ (binary or labeled). Top: 3C-LSTM-UNET. Network architecture was implemented for the SNEMI3D dataset to optimize for accuracy. The inputs are processed with three consecutive Conv-LSTM modules, followed by a symmetric Residual U-Net structure. Bottom: 3C-Maxout. Network architecture was implemented for the Harvard Rodent Cortex and PNS datasets to optimize for speed.}
\label{Fig: LSTM_unet and maxout structure}
\end{figure*}

\textbf{3-D reconstruction system:} Our 3-D reconstruction system 
consists of the following steps (shown in  Figure~\ref{fig:TopView}):

{\bf 1)}
Seeding and labeling the volume with an initial imperfect 2-D/3-D instance segmentation that overlaps all objects except for their boundaries (over-segmentation). 

{\bf 2)} Encoding the labeled seeds into a new space using the 3C rule $\chi$.

{\bf 3)} Applying a fully convolutional network $\log(N)$ times to transfer the projected labeled seeds from the source image to the target images, and then take their cross labeling. 

{\bf 4)} Decoding the set of network outputs to the original label space using the inverse 3C rule $\chi^{-1}$. 

{\bf 5)} Agglomerating the labels into 3-D consistent objects based on the overlap of the original seeding and the segments predicted from other sections. 

To initially seed and label the volume (Step 1), 
we compute and label 2-D masks that over-segment all objects. For this we follow common practice in connectomics, computing object borders with FCNs and searching for local minima in 2-D on the border elevation maps. Subsequently (Step 2), 
we use $\chi$ to encode the seeds of each section, resulting in a $k$-tuple over the $l$-color alphabet for each seed ($k{=}5$ and $l{=}4$ in Figure~\ref{fig:relabeling_for_Cross_Classification}).
 
A fully convolutional neural network then predicts the correct label mapping between interrelated images of sections $Z$ and $Z{\pm}W$, which determines which pixels in target image $Z{\pm}W$ belong to which seeds in source image $Z$ (step 3). All seeds here are represented by a fixed number $l$ of colors, and prediction is done $\log(N)$ times based on Equation \ref{eq:encoding}. For decoding, all $\log(N)$ predictions are aggregated for each pixel to determine the original label of the seed using $\chi^{-1}$ (Step 4). For training, we use saturated ground truth of the 3-D consistent objects. This approach allows us to formalize the reconstruction problem as a tracking problem between independent images, and to deal with the tendency of objects to disappear/appear in different portions of an enormous 3-D dataset.
 
 
We now describe how the 3C seed transfers are utilized to agglomerate the 2-D masks (as shown in Figure~\ref{fig:global edge merge}). For agglomeration (Step 5), the FCN for 3C is applied from all source images to their target images, which are at most $W$ image sections apart from each other across the image stack (along the $z$ dimension). We collect overlaps between all co-occurring segments, namely, those occurring by the original 2-D seeding, and those by the 3C seed transfer from source to target images. This leaves $2W{+}1$ instance segmentation cases for each image (including the initial seeding), which directly link seeds of different sections. Formally, the overlaps of different labels define a graph whose nodes are the 2-D seed mask labels and the directed weighted edges are their overlap ratio from the source to the target. Instead of optimizing this structure (as in the {\it Fusion} approach of \cite{kaynig2015large}), we found that agglomerating all masks of sufficient overlap delivers adequate accuracy even for a small $W$. We do however make forced linking on lower probability edges to avoid ``orphan'' objects that are too small, which is biologically implausible. We provide further details in the Supplementary Materials.

We note that 3C does not attempt to correct potential merge errors in the initial seeding. These can be addressed post-hoc by learning morphological features \cite{rolnick2017morphological,zung2017error} or global constraints \cite{brianHarvardCVPR2019}.

\textbf{Adaptation to other domains:} To leverage 3C for multi-object tracking in videos, a domain-specific seeder should precede cross classification (e.g. with deep coloring \cite{kulikov2018instance}). Natural images are likely to introduce spatially consistent object splits across frames and hence a dedicated agglomerating procedure should follow. The 3C technique can be readily applied to other medical imaging tasks, with seed transfers across different axes for isotropic settings.

\section{Experiments}
The SNEMI3D challenge is a widely used benchmark for connectomic segmentation algorithms dealing with anisotropic EM image stacks~\cite{arganda2015crowdsourcing}. Although the competition ended in 2014, several leading labs recently submitted new results on this dataset, improving the state-of-the-art. Recently Plaza et al.~suggested that benchmarking connectomics accuracy on small datasets as SNEMI3D is misleading as large-scale ``catastrophic errors'' are hard to assess~\cite{PlazaBerg2016,plaza2018analyzing}. Moreover, the clustering metrics such as Variation of Information~\cite{meilua2007comparing} and Rand Error~\cite{unnikrishnan2007toward} are inappropriate since they are not centered around the connectomics goal of unraveling neuron shape and inter-neuron connectivity. We therefore conduct experiments on three additional datasets and show the Rand-Error results only on the canonical SNEMI3D dataset. 
To assess the quality of 3C at large scale, we demonstrate results on the widely studied dataset by Kasthuri et al.~\cite{kasthuri2015saturated} ({\it S1 Dataset}). To further assess our results in terms of the end-goal of connectomics, neuronal connectivity, we evaluate the synaptic connectivity of the 3C objects using the NRI metric~\cite{reilly} ({\it ECS Dataset}).
In the final experiment we focus on the tracking ability of 3C 
({\it PNS Dataset}). 
 
\begin{figure*}[t] 
\centering
	 \includegraphics[width=\linewidth]{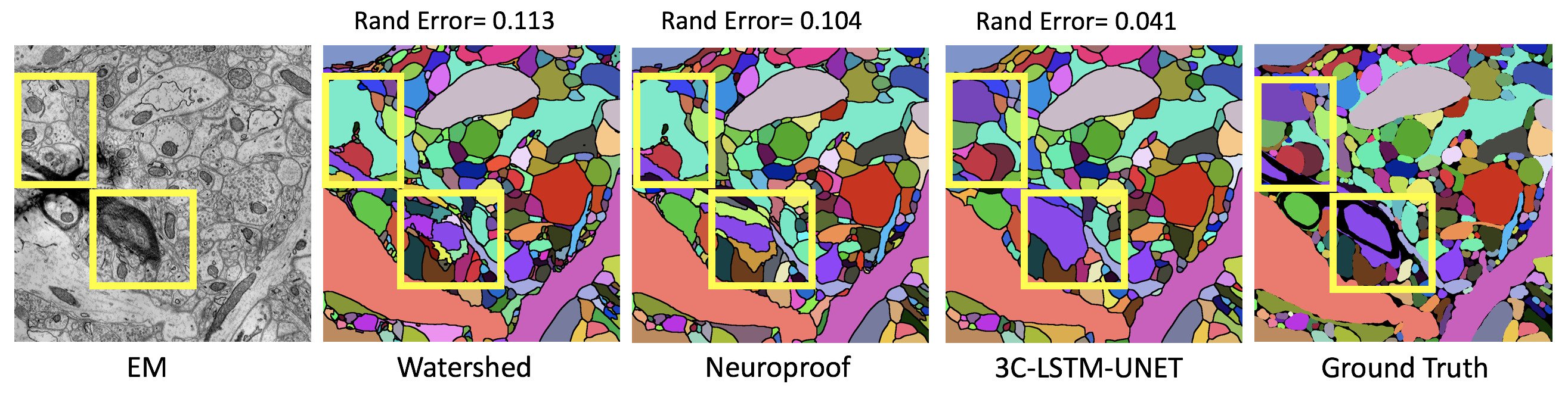} 
	 \caption{SNEMI3D: The 3C-LSTM-UNET Results compared with baseline techniques: Watershed, Neuroproof, and ground truth. }	
	 \label{fig:compared with baselines}
\end{figure*}

\subsection{SNEMI3D Dataset}
\label{sec:SNEMI3D}
In order to implement 3C on the SNEMI3D dataset, we first created an initial set of 2-D labeled seeds over the entire volume. These were generated based on the regional 2-D minima of the border probability map. This map was generated by a Residual U-Net, which is known for its excellent average pixel accuracy in border detection~\cite{lee2017superhuman,UNET}. Next, the 3C algorithm was used to transfer 2-D labeled seeds through the volume, as shown in Figure~\ref{fig:relabeling_for_Cross_Classification}. Finally, the original 2-D labeled seeds and transferred labeled seeds were agglomerated if their overlap ratio exceeded 0.1. We found that $W{=}2$ delivers adequate accuracy. All orphans were greedily agglomerated to their best match. In order to achieve better accuracy, we tested 3C with various network architectures, and evaluated their accuracy. To date, convolutional LSTMs (ConvLSTM) have shown good performance for sequential image data~\cite{xingjian2015convolutional}. In order to adapt these methods to the high pixel-accuracy required for connectomics, we combined both ConvLSTM and U-Net.
The network is trained to learn an instance segmentation of one image based on the proposed instance segmentation of a nearby image with similar context. 
We found that the LSTM-UNET architecture has validation accuracy of 0.961, which outperforms other commonly used architectures. 
A schematic view of our architecture is given in Figure~\ref{Fig: LSTM_unet and maxout structure}. Details are provided in the Supplementary Materials. 

In order to illustrate the accuracy of 3C, we submitted our result to the public SNEMI3D challenge website alongside 
two common baseline models, the 3-D watershed transform (a region-growing technique) and Neuroproof agglomeration~\cite{parag2015context}. Our watershed code was adopted from \cite{Matveev2017Multicore}. Similar to other traditional agglomerating-techniques, Neuroproof trains a random forest on merge decisions of neighboring objects 
\cite{nunez2013machine,parag2015context,parag2017anisotropic}.   
These baseline methods were fed with the same high-quality border maps used in our 3C reconstruction system. The comparisons of 2-D results with ground truth  (section $Z{=}30$) are shown in Figure~\ref{fig:compared with baselines}. Our result has fewer merge- and split-errors, and outperforms the two baselines by a large margin. Furthermore, 3C compares favorably to other state of art works recently published in Nature Methods~\cite{beier2017multicut,FFN}. In the SNEMI3D challenge leaderboard 
the Rand-Error of 3C was 0.041, compared with the 0.06 achieved by a human annotator. Our accuracy (ranked 3rd) outperforms most of the traditional pipelines 
many by a large margin, and is slightly behind the slower neuron-by-neuron 
FFN segmentation for this volume. 
The leading entry is a UNET-based model learning short and long range affinities \cite{lee2017superhuman}.
The results are summarized in Table~\ref{Tab:compared with baselines}.

\begin{table}
\begin{center}
\begin{tabular}{|c|c|c|c|c|c|} 
\hline 
Model & Rand & VI  &$\text{VI}_\text{split}$ & $\text{VI}_\text{merge}$ & Complexity \\
\hline\hline
%
Watershed     &0.113 &0.67  &0.55  &0.12  &- \\
Neuroproof     &0.104 &0.55  &0.42 &0.13  &- \\
Multicuts
&0.068 &0.41 &0.34 &0.07 &- \\
\textbf{3C} &\textbf{0.041}  &\textbf{0.31} &\textbf{0.19} &\textbf{0.12} &\textbf{
O($V \log N $)}\\
FFN
&0.029 & - &- &- & O($V  N$)\\
Human val. &0.060 &- &- &- &-\\
\hline
\end{tabular}
\end{center}
\caption{Comparison of Watershed, Neuroproof \cite{parag2015context}, Multicut \cite{beier2017multicut}, human values, 3C and FFN \cite{FFN}
on the SNEMI3D dataset for Rand-Error, Variation of Information
{\it VI}, VI split, VI merge. Time Complexity: $N$ is the number of objects and $V$ is 
number of pixels. For empirical comparison see the performance section. We do not have access to the FFN and human outputs and hence their VI metric is missing.}



\label{Tab:compared with baselines}
\end{table}

\subsection{Harvard Rodent Cortex Datasets ({\it ECS}, {\it S1})} 
\begin{figure} 
\centering  
	 \includegraphics[width=0.85\linewidth]{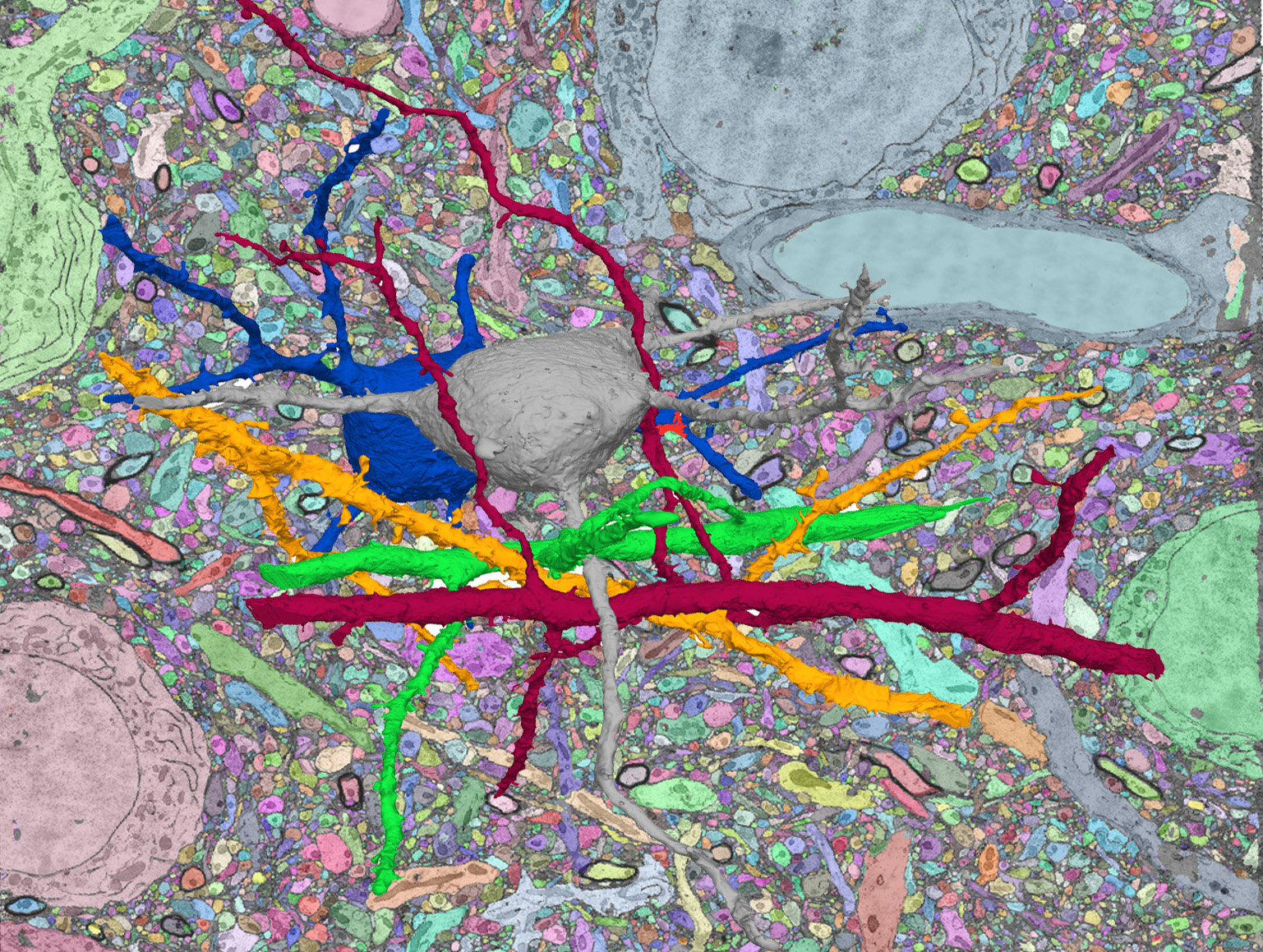} 
	 \caption{Results on Kasthuri et al. \cite{kasthuri2015saturated} {\it S1}. Fast lightweight 3C-Maxout operating on 3-D seeds, without agglomeration. Background: Segmented section. Foreground: five 3-D reconstructed objects.     }
	 \label{fig:voxelnet-5-objects}
\end{figure}
We describe two additional tests: (1) 3C on datasets with known synaptic connectivity (subsets of {\it ECS} and {\it S1}), and (2) a lightweight agglomeration-free reconstruction applied to a large-scale dataset ({\it S1}). 

\textbf{Connectivity-based test:} Following \cite{plaza2018analyzing}, which recently advocated connectivity-based evaluation of connectomics, the accuracy of the pipeline was evaluated using the NRI metric~\cite{reilly}. In a nutshell, the NRI ranges between $0$ and $1$, measuring how well a given neuronal segmentation preserves the object connectivity between neural synapses ($1$ being optimal). 

For the first test, we used a lightweight yet successful FCN model \cite{Matveev2017Multicore} ({\it Maxout}) (for border and 3C computations), reconstructing the test set of \cite{kasthuri2015saturated} ({\it S1}) (3C with FOV of 109 pixels). Maxout is currently the fastest border detector in connectomics, which was previously successfully used for single-object tracking~\cite{meirovitch2016multi}. Details of architecture and training are presented in the Supplementary Materials.

The NRI score of the 3C-Maxout segmentation was 0.54, compared to 0.41 of a traditional agglomeration pipeline \cite{Matveev2017Multicore}. For the second test, we were granted permission to reconstruct a recently collected rat cortex dataset of the Lichtman group at Harvard ({\it ECS}). This test allowed the comparison of 3C to the excellent agglomerative approach of \cite{parag2017anisotropic} (4th on SNEMI3D), while using exactly their U-Net~\cite{UNET} border predictions as inputs to our 3C network. On the test set our NRI score was 0.86, compared to 0.73 for the agglomeration pipeline. 

\textbf{Large-scale reconstruction ({\it S1}):} We also ran a fast version of 3C on the entire S1 dataset (90 gigavoxels: 
1840 slices, 6 nm x 6 nm x 30 nm per voxel). In this experiment, we omitted the agglomeration step of the reconstruction algorithm to achieve better scalability and let 3C run on 3-D masks computed by local minima of the border probability maps. This  implementation is highly scalable since it has no agglomeration step, while the 3C masks are updated on-the-fly in a streaming fashion every 100 slices. The Maxout implementation is attractive for large-scale systems because it is efficiently parallelized on multi-core systems with excellent cache behavior on CPUs \cite{Matveev2017Multicore}. 
Figure \ref{fig:voxelnet-5-objects} shows five objects that span the whole volume of S1.
%
%

\subsection{Peripheral Nervous System ({\it PNS}) Dataset}
\begin{figure} 
\centering
	 \includegraphics[width=0.9\linewidth]{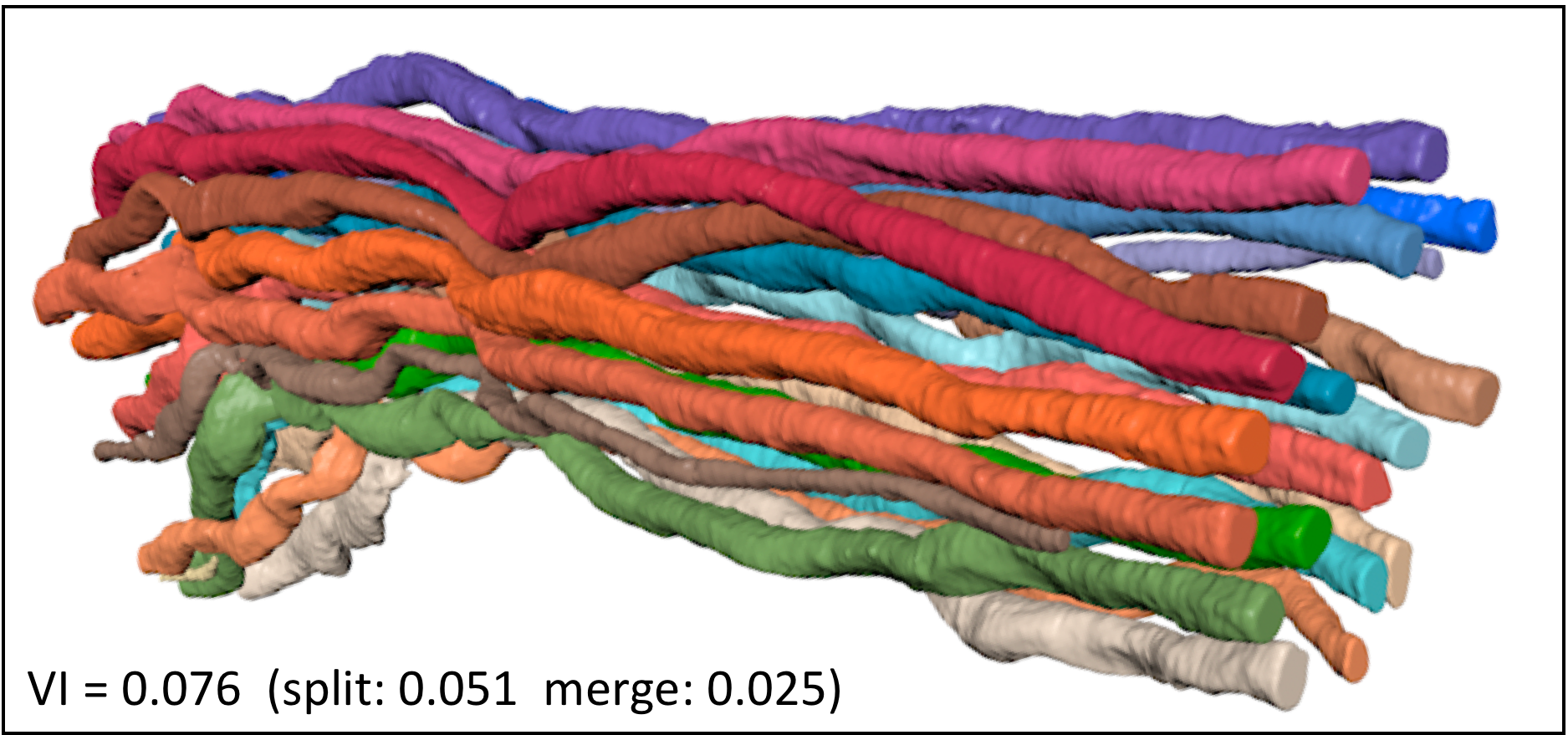} 
	 \caption{3C-Maxout results of recursively tracking all objects (axons) directly from the PNS raw images (no post-processing). }	
	 \label{fig:pns-datasets}
\end{figure}
Next, we tested the ability of the 3C framework to track objects recursively based on raw images in a streaming mode, that is, independently of any agglomerating or post-processing steps.
 

We chose a previously unpublished motor nerve bundle from a (newborn) mouse contributed by the Lichtman lab at Harvard for this purpose because it is a closed system in which all objects are visible in the first and last image sections of the 915-images dataset. This dataset is important to neurobiologists since it contains the entire neural input (21 axons) of a complete muscle. 

Again, we applied the 3C algorithm using the lightweight FCN Maxout architecture of  \cite{Matveev2017Multicore}. 
3C was able to track all objects without erroneous merges; results are shown in Figure~\ref{fig:pns-datasets}. Out of the 21 axons, 20 were recursively reconstructed to their full extent (split errors in only one object). One extremely thin axon disappeared from the image and reappeared after 7 sections and was not reconstructed. 
The axon run-length for all reconstructed axons was above 70 microns (and $> 900$ sections) until all of these exited the volume on the last slice in the image stack. 


This benchmark demonstrates: {\bf a)} 3C-Maxout performs well in a tracking task directly from raw images, even in the difficult connectomics regime, and {\bf b)} our training procedures display satisfactory generalization abilities, learning from a relatively small number of examples. 

\section{Scalability Comparisons}
\label{sec:performence}


\begin{figure}
\centering
	 \includegraphics[width=1.0\linewidth]{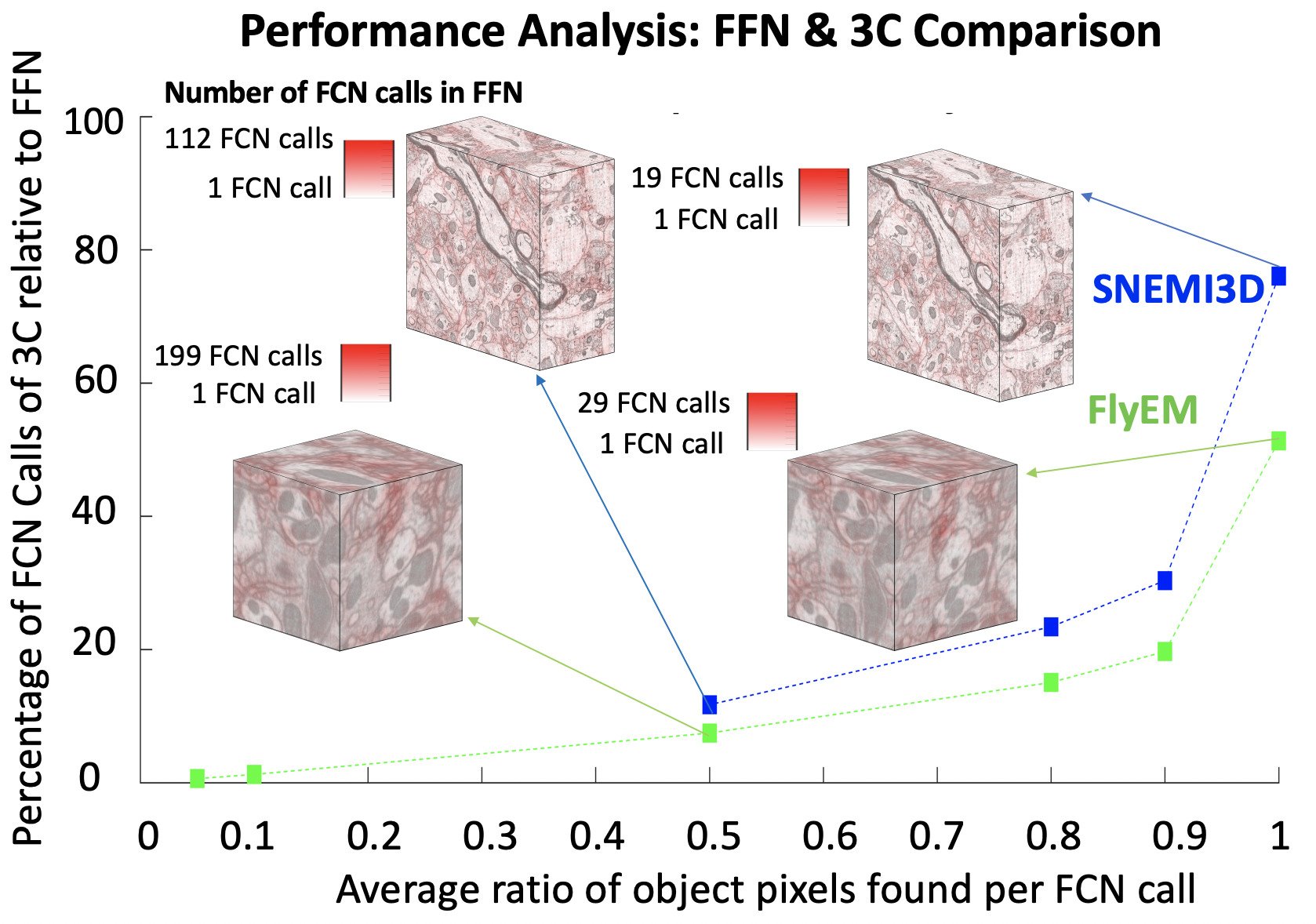} 
	 \caption{
	 Compute cost per pixel using FFN-Style segmentation. We computed the number of times a pixel is participating in object detection (red) for two public datasets (SNEMI3D, FlyEM), and compared to the number of classification calls in 3C. 
	 %
	 }	
	 \label{Fig: FFNEmpiricalCostPerPixel}
\end{figure}


In this section, we compare the relative scalability of 3C to FFN and MaskExtend, as far as possible without having access to the full FFN pipeline. 
3C is a generalization of the FFN and MaskExtend techniques~\cite{FFN,meirovitch2016multi}, which augment the (pixel) support set of each object, one at a time. The 3C technique simultaneously augments all the objects from its input image(s) after a logarithmic number of iterations (see Figure \ref{fig:relabeling_for_Cross_Classification}). This allows us directly to compare the two types of approaches based on the number of iterations required, ignoring details of implementation.

\textbf{FFN: } We compare the number of FCNs calls in FFN and 3C assuming both algorithms reconstruct all objects flawlessly. We assume both algorithms use the same FCN model. Although 3C and FFN invoke FCNs a logarithmic versus a linear number of times, respectively, FFN runs on smaller inputs, centered around small regions of interest. 

At each iteration, FFN will output an entire 3-D mask of the object around the center pixel. We assume that a fraction of those pixels will require revisiting (zero for best-case scenario). Figure \ref{Fig: FFNEmpiricalCostPerPixel} depicts the number of FCN calls and their ratio for FlyEM~\cite{takemura2015synaptic} and SNEMI3D~\cite{arganda2015crowdsourcing} for FFN and 3C. In 3C, each pixel participates in an FCN call a number of times logarithmic in the number of objects visible in the field of view (the FCN calls for FFN are color-coded red in the data cubes of Figure \ref{Fig: FFNEmpiricalCostPerPixel}). The $y$-axis depicts the ratio between the FCN calls by 3C to that for FFN, for several ratios of object pixels found per FCN call. A zero ratio means that no pixels are found for the object in a single FCN call, whereas 1 means that all object pixels are found and require no further revisiting. We can see from the plot that, assuming error-free reconstruction, 3C is more efficient than FFN when there is a fraction of object pixels that require revisiting after a single call of the FCN. The revisiting of some pixels is also reported by~\cite{FFN}, as the 3-D output has greater uncertainty far from the initial pixels. For a revisit ratio of 0.5, 3C is more than 10x faster than FFN on FlyEM. 
  
\begin{figure} 
\centering
	 \includegraphics[width=.9\linewidth]{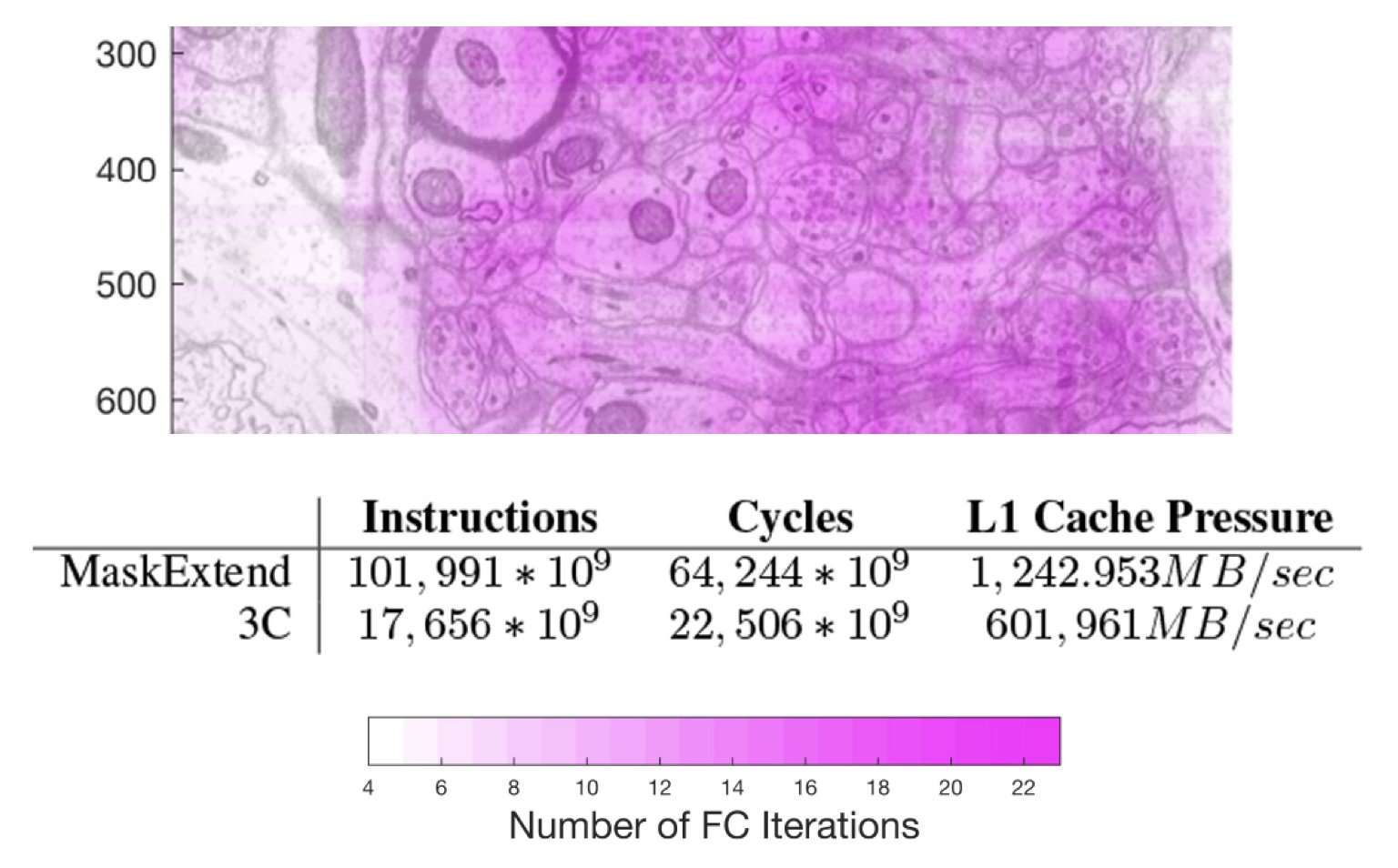} 
	 \caption{Compute cost per pixel with single-object tracking methods~\cite{meirovitch2016multi}. The number of calls per pixel is color-coded in purple. For the highly dense areas 23 calls of the object detector are required. The table depicts the performance counter statistics for the execution of \cite{meirovitch2016multi} and the 3C-Maxout FCN on a stack of 100 images.}	
	 \label{Fig: MaskExtendEmpiricalCostPerPixel}
\end{figure}

\textbf{MaskExtend:} 
Figure~\ref{Fig: MaskExtendEmpiricalCostPerPixel} repeats the above procedure with MaskExtend~\cite{meirovitch2016multi}, comparing its FCN calls with 3C. MaskExtend is more wasteful than 3C, propagating some pixels into its FCN model 23 times. The instruction and cycle counts as well as the L1 Cache pressure are larger for MaskExtend (equal multi-core infrastructure and inference framework~\cite{Matveev2017Multicore}).

\section{Conclusion} 
\label{sec:discussion and future work}

In this paper, we have presented cross-classification clustering (3C), an algorithm that tracks multiple objects simultaneously, transferring a segmentation from one image to the next by composing simpler segmentations. We have demonstrated the power of 3C in the domain of connectomics, which presents an especially difficult task for image segmentation.
Within the space of connectomics algorithms, 3C provides an end-to-end approach with fewer ``moving parts,''  improving on the accuracy of many leading connectomics systems. Our solution is computationally cheap, can be achieved with lightweight FCNs, and is at least an order of magnitude faster than its relative, flood-filling networks. Although the main theme of this paper was tackling neuronal reconstruction, 
our approach also promises scalable, effective algorithms for broader applications in 
medical imaging and video tracking.

 
\section*{Acknowledgements} 
\label{sec:acknowledgements} 
We would like to thank Jeff Lichtman and Kai Kang for
allowing us to access the PNS dataset, Marco Badwal for alignment, and Daniel Berger and Casimir Wierzynski for insightful comments. 
This research was supported by the National Science Foundation (NSF) under grants IIS-1607189, IIS-1447786, CCF-1563880, IIS-1803547 and by a grant from the Intel corporation.
 
{\small
\bibliographystyle{ieee_fullname}
\bibliography{reconstruction}
}

\end{document}